# Depth-based Selective Blurring in Stereo Images Using Accelerated Framework

Subhayan Mukherjee

*National Institute of Technology Karnataka,*
*Surathkal, Mangalore, India.*
+91-9742138607
subhayan001@gmail.com

Ram Mohana Reddy Guddeti

*National Institute of Technology Karnataka,*
*Surathkal, Mangalore, India.*
+91-824-2473551
profgrmreddy@nitk.ac.in

**Abstract**: We propose a method for depth-based selective blurring of non-interest regions of stereo images using CPU-GPU accelerated framework based on Java Thread Pool and APARAPI. Our method combines both block and region-based stereo matching approaches and generates depth maps using disparity measurements of only 18% image pixels (left or right). The methodology involves segmenting pixel lightness values using fast K-Means implementation, refining the segment boundaries using morphological filtering and connected components analysis; then determining the boundaries' disparities using SAD cost function. Complete disparity maps are reconstructed from these boundaries' disparities. Gaussian blur is applied to de-focus image regions lying in depth ranges of user's non-interest. Experimental results from Middlebury dataset show that our method outperforms traditional disparity estimation approaches using SAD and NCC by up to 33.6% and some recent methods by up to 6.1%. Further, our acceleration framework achieves speed-up of 5.8 for 250 stereo video frames (4096x2304 resolution).

## Introduction

Stereo vision based disparity calculation is an important research problem in computer vision with applications such as robotic vision, 3D scene reconstruction, object detection and tracking, etc. [2]. Here, the main challenge is to obtain an accurate depth information of a scene by comparing the pixels of left and right image of that scene. It is challenging because individual pixels contain only the colour and spatial information and represent the low level image features [3]. So, to effectively compare a stereo image pair, we need to develop the framework to identify appropriate high level features and thereby comparing these features efficiently with reasonable accuracy and speed.

Disparity of a pixel varies inversely as the distance of a point in a scene (that the pixel represents) from the 3D camera. A disparity or depth map contains the mapping of each pixel of an image to its corresponding disparity. A cost function is used to quantify the similarity between pixels of the left and right images of a stereo image pair. Then, pixels' disparities are estimated by relative displacement of the corresponding matching counterparts across the image pair.

For finding the matching pixels between the stereo image pair, we have two basic approaches: block-based and region-based; the strengths and weaknesses of these two basic approaches are summarized in Table 1 [1]. The key challenge in block-based methods is the determination of an optimal block size, and in region-based methods is the detection of gradually changing disparities. In estimating the stereo disparities using block-based methods, small block sizes produce sharp edges on the depth map but generates errors in the homogeneous regions since only a very limited local information is utilized. On the other hand, large block sizes provide good performance in homogeneous areas but yield very inaccurate disparity measurements along objects' edges. This is because, pixels inside the same block may have varying disparities. For region-based methods, apart from the fact that they not only create a limited number of depth levels but also require accurate results of segmentation, which makes them inherently computationally intensive and time-consuming. This motivates us to propose a novel hybrid method by combining these block-based and region-based approaches and thereby overcoming their individual limitations.

Table 1  Comparison of Basic Depth Estimation Approaches

|  | ***Block-based Method*** | ***Region-based Method*** |
| --- | --- | --- |
| ***Approach*** | Depth estimation based on information contained in pixels and their surroundings. | Depth estimation based on optimal value of cost function for entire image regions of pixels with similar disparities. |
| ***Strength*** | High resolution depth maps. | Sharp edges on depth maps. |
| ***Weakness*** | Need to determine optimal block sizes for different images, or even different regions of the same image for creating accurate depth maps. | Unsuitable for finding the disparities that are gradually changing, as all pixels constitute even a large region share a constant disparity. |

To overcome these problems, we propose a method of disparity estimation by segmenting the pixels' lightness values by a fast histogram-based K-Means implementation and refinement of segment boundaries using morphological filters and connected components analysis.

The core idea behind our proposed method is to use a scalable, block-based approach to estimate the disparities of only pixels lying on the refined segment boundaries, taking large block sizes to get reliable disparity estimates in less number of computations. So, our method is scalable to high resolution stereo image pairs. Then, the proposed disparity map reconstruction method is used for estimating the disparities of pixels lying inside segment boundaries and the details are given in Section III. Finally, a Gaussian kernel is used to blur image regions corresponding to depth ranges of user's non-interest. Thus, proposed methodology for the selective blurring phase is similar to the methodology of [10]; but our method is not at all concerned with the occlusion handling, however the proposed scheme can handle multiple, and discrete depth ranges of interest.

Further, our method utilizes two-dimensional intensity based segmentation of the left image whereas authors of [1] used a one-dimensional colour-based segmentation process of individual rows of pixels. Similarly, our proposed method uses only 'L' values of pixels, making the process simple and fast; on the other hand the method of [3] is based on object-based segmentation using colour, spatial and shape information.

Lastly, we demonstrate how the running time of our proposed algorithm can be reduced by running some of its mutually independent operations in parallel on the multiple cores of CPUs and GPUs. We also experiment with combining the power of the CPU and the GPU to achieve even higher degrees of parallelism.

Key contributions of the proposed work are as follows:

- To the best of our knowledge, this is the first work on a hybrid method for selective blurring of stereo images using disparity information obtained by segmentation of only lightness values of left image pixels, unlike other methods using colour, texture and shape characteristics.
- Further, to the best of our knowledge, this is the first paper dealing with morphological filters and connected component analysis to get sparse, but accurate disparity estimates used for selective blurring of stereo images.
- Lastly, to the best of our knowledge, this is the first work which proposes an accelerated framework for reducing the running time of a novel algorithm for selective depth-based blurring, using Java Thread Pool (CPU) and APARAPI (GPU), and further, combines them to achieve even greater parallelism.

The rest of this paper is organized as follows: Section II deals with Related Work; Section III explains our Proposed Algorithm with both Sequential and Parallel Time Complexity Analysis; Section IV gives the Results and Discussion; Finally, Section V concludes with future directions.

# Related Work

Most existing stereo disparity algorithms use one of the two types of measures of pixel similarity between left and right image: pixel-to-pixel or window-based. Pixel-to-pixel algorithms find the pixel similarity measure solely on individual pixel values. Examples are Absolute Difference (AD) or Squared Difference (SD). On the other hand, window-based methods aggregate the pixel matching cost over a support region around a pixel of interest, e.g. Sum of Absolute Differences (SAD) or Sum of Squared Differences (SSD) [12]. A study on comparison of cost functions commonly used in recent methods of stereo disparity determination [13] has come to the conclusion that the performance of a matching cost function depends on the stereo method that uses it. The cost aggregation strategies used in recent stereo disparity estimation methods can be classified into cost aggregation based on rectangular windows-based and unconstrained shapes-based strategies, and lastly, and by assigning different and variable weights to pixels falling in the neighbourhood of the two points on the reference and target images for which the stereo correspondence is being evaluated. Rectangular window-based methods can be further classified into variable window-size and/or offset-based, multiple-windows-based, and differential weights of window points-based approaches [14].

To determine the best matching pixel for a given pixel, the most obvious means involves selecting the point in the other image within a certain disparity range that has the best similarity value, and this has been used in many stereo matching methods. This strategy is called 'Winner Takes it All' (WTA). However, a stereo algorithm based only on WTA does not consider parts of stereo images that are only visible in one of the two images, called occlusions. Moreover, image regions having little or repetitive texture yield similar matching costs. In these areas, WTA is error-prone as there isn't a clear optimum correspondence.

The smoothness constraint for stereo states that disparity does not change much on object surfaces. This concept is often used to improve disparity estimates. If a correct disparity has been found then it can be used to constrain the range of possible disparities of other points on the same surface. We incorporated this concept to reduce wrong estimates on areas with little or repetitive texture.

Morphological filters were used for partitioning a low depth-of-field image into focused object of interest and blurred background [15]; but, our proposed method uses morphological filters for an entirely different task altogether, viz. refining segment boundaries whose disparities are determined later to aid in the process of synthesizing a scene with only users' depths of interests in focus and everything else blurred, which mimics a low depth-of-field image.

Connected components analysis based methods have aided image segmentation for the purpose of text detection [16-18], brain tissue analysis [19, 20] and colour image segmentation using genetic algorithms [21], whereas in our proposed work, we use connected components analysis to refine the boundary map by removing small, isolated connected components (artefacts) which may yield incorrect depth estimates. Our approach may seem similar to [19], where the authors first roughly distinguish the region of non-brain tissue through connected component labelling, and then try to refine those edges using the morphological operations, dilation and erosion. However, a slightly closer look at our method would reveal the fact, that we perform the connected components analysis *after* applying the morphological filters and thus the ordering of these two operations is entirely reversed.

Several depth estimation techniques [4-7] have been developed in recent years to overcome the limitations of two basic approaches of stereo disparity estimation, viz. block-based and region-based. Their key features are summarized in Table 2.

**Table 2** Comparison of Recent Depth Estimation Methods

| *Work* | *Advantages* | *Limitations* |
|---|---|---|
| Choi [4] | Bi-directional consistency check of disparity blocks can improve disparity estimates. | Comparison with similar algorithms using a standard dataset has not been done. |
| Zhu and Yu [5] | Self-adaptive block matching can increase the efficiency of searching, and reduce errors in matching and block-shape. | Presence of noise in the original left or right images can result in selection of incorrect size for block-matching (8x8 or 16x16). |
| Wang and Zheng [6] | Cooperative optimization can check or fix disparity errors. | Uses the time-consuming mean-shift algorithm for segmentation. |
| Lu and Du [7] | Harris corner point extraction and feature-matching yields better corresponding points than traditional methods. | Both left and right images are scanned for corner extraction and matching in the initial steps, making them time-consuming. |

## Proposed Algorithm

Our proposed algorithm obtains the depth map as per the flowchart given in Fig. 1. The steps are explained below.

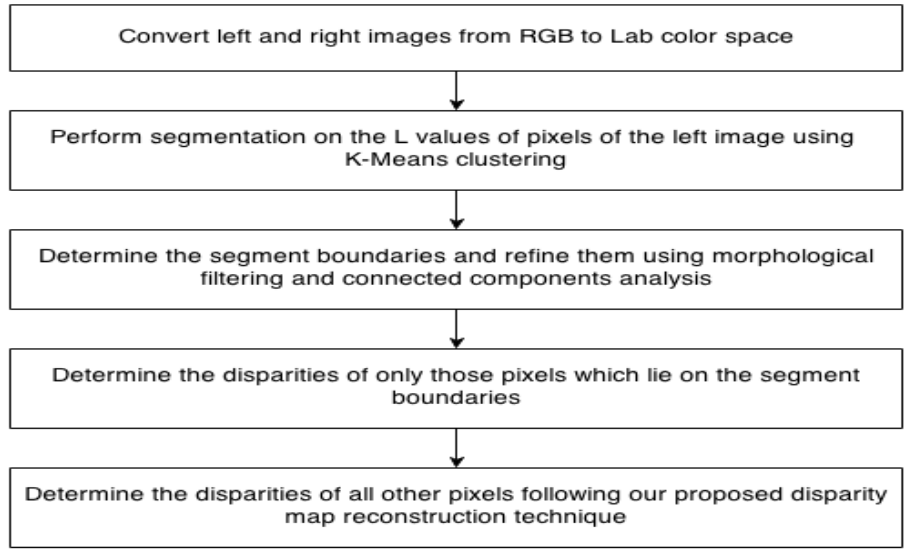

**Fig. 1** Flowchart of Our Proposed Depth Extraction Algorithm

**Colour Space Conversion:**

Human eyes are more sensitive to changes in brightness than in colour. The 'L' component of the Lab colour space closely matches the human perception of lightness. Hence, by applying the pertinent transformations discussed in [9], we convert the left and right images from RGB to the Lab colour space and retain only the 'L' values of its pixels.

**Segmentation:**

'L' values of the left image pixels are segmented using a fast implementation of the K-Means clustering algorithm. We build a histogram of the 'L' values and use that histogram instead of the actual pixel values for clustering. Thus, the runtime of clustering is significantly reduced, as we perform clustering on a small, fixed number of bins comprising the histogram. Since there exists a one-to-one correspondence between each pixel and the bin to which it has been mapped, we can easily identify the cluster to which the pixel has been assigned as the cluster to which its bin has been assigned.

**Segment Boundary Detection and Refinement:**

Segment boundary detection is achieved by comparing the cluster assignment of each pixel with that of its 8-connected pixels (i.e. its Moore neighbourhood). If any of them is found to differ, we mark the pixel as '1' (falling on a segment boundary), else as '0' (not on a segment boundary). So, this step creates the "boundary map" after segmentation.

But this approach also falsely identifies many pixels as belonging to segment boundaries due to limitations imposed by clustering accuracy, as we segment the image based on only the pixels' lightness (L) values and not on their colour components (a, b) or spatial locations (x, y) in the image. So, we apply two morphological filters to refine the boundary map by removing such noisy pixels, in the following order:

Fill: Fills isolated interior pixels such as the centre pixel in:

$$\begin{matrix} 1 & 1 & 1 \\ 1 & 0 & 1 \\ 1 & 1 & 1 \end{matrix}$$

Remove: Removes interior pixels, i.e., sets a pixel to '0' if all its 4-connected neighbours are '1', thus leaving only the boundary pixels on.

Further, we use connected components analysis and remove small artefacts in the boundary map due to segmentation errors, by ordering connected components present in the boundary map by the number of pixels constituting each connected component to remove the smallest connected components which contribute about 4% of the total number of boundary pixels.

**Disparity Measurement of Boundary Pixels:**

We assume that the left and right cameras are calibrated and the left and right images share the same image plane. So, the correspondence of each pixel can only be in the horizontal direction. For e.g., the correspondent point of any pixel on the left image can only appear on the same row of the right image.

We use the SAD (Sum of Absolute Differences) [2] cost function to determine only the disparities of boundary pixels, using the 'L' values of the left and right image pixels. It should also be noted here, that by "boundary" pixels, we are also (implicitly) referring to the pixels of the left image that map to the left and right borders of the disparity map.

**Disparity Map Reconstruction from Boundaries:**

Our disparity map reconstruction algorithm scans through each row of the partially computed disparity map and computes the remaining disparities based on disparities that have already been calculated. It operates in two stages:

***Disparity Propagation ('Fill' Stage):***

In the first stage, we scan the disparity map row-wise, left to right—whenever we consecutively encounter two boundary pixels with equal disparity values, we 'fill' the intermediate pixels with that disparity value. This reflects our assumption that the pair of points in consideration actually belong to the same object in the original image. So, all their intermediate pixels also belong to that same object, and hence, should have similar disparity values. The process is explained using the pseudo-code below; *disp_map* is the matrix containing the disparities of boundary pixels and disparity(*cell*) refers to the disparity value of a boundary pixel of *disp_map*.

**Algorithm 1** Disparity Propagation along Scan Lines

1: **for** each *row* of *disp_map*, starting from its *top* **do**
2:     **for** each *cell* of this *row*, starting from its *left* **do**
3:         **if** this *cell* is a pixel on a segment boundary, **then**
4:             **if** disparity(*cell*) == disparity(previous boundary pixel), **then**
5:                 disparity of all intermediate pixels ← disparity(*cell*)
6:             **end**
7:         **end**
8:     **end**
9: **end**

*Estimation from Known Disparities ('Peek' Stage):*

In the second stage, for all pixels whose disparities have not yet been determined, we estimate their disparities by 'peek'-ing at disparity values of their two nearest pixels for which the disparities have been already determined. We search for these two nearest pixels along the same column as the pixel in question and the details are as follows:

**Algorithm 2** Disparity Estimation from Known Disparities

1: **for** each *row* of *disp_map*, starting from its *top* **do**
2:     **for** each *cell* of this *row*, starting from its *left* **do**
3:         **if** disparity of this *cell* has not been already determined, **then**
4:             *disp_n* ← {disparities of two nearest pixels on the same column}
5:             *disp_r* ← statistical range of *disp_n*
6:             **if** *disp_r* > *disparity_threshold*, **then**
7:                 disparity of this *cell* ← smaller of the two values in *disp_n*
8:             **else**

9:          disparity of this *cell* ← mean of the two values in *disp_n*
10:        **end**
11:     **end**
12:   **end**
13: **end**

*Depth-based Selective Blurring:*

The depth map output by the aforementioned final step is fed to the blur map generator, along with the depths of users' interest and a overall blur level. Thus, the blur map determines which pixels in the image will be blurred. The blur map is fed to a depth-based blurring module, which synthesises a new scene, where all depth ranges of users' non-interest are blurred using Gaussian function $h_g(n_1, n_2)$ and Blurring kernel $h(n_1, n_2)$ as defined by Eq. 1 and Eq. 2 respectively,

$$h_g(n_1, n_2) = e^{\frac{-(n_1^2 + n_2^2)}{2\sigma^2}} \quad (1)$$

$$h(n_1, n_2) = \frac{h_g(n_1, n_2)}{\sum_{n_1}\sum_{n_2} h_g} \quad (2)$$

where $n_1$ and $n_2$ are dimensions of the blurring kernel, $\sigma^2$ is the variance for building the Gaussian kernel and the overall blur level is determined by $\sigma$. Fig. 2 shows the overall process of this depth-based selective blurring.

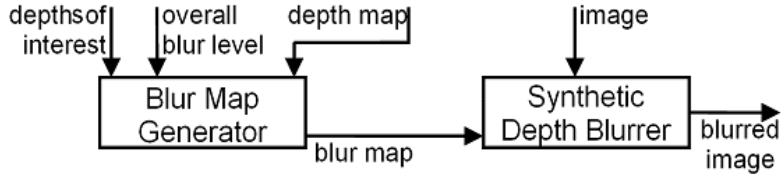

**Fig. 2** Depth-based Selective Blurring

*Sequential Complexity Analysis:*

Here, we present the step-by-step time complexity analysis of our proposed algorithm and thereby identifying computationally intensive steps and the degrees of interdependency between the operations of each step. Next, we convert steps eligible for data-parallelization to data-parallel workloads in JTP and APARAPI.

Our first step involves colour space conversion, in which we apply predefined transformations [9] to the (R, G, B) triplet of each pixel in both the left and right image to extract the 'L' values from them. Thus, the time complexity of this step is $O(c_1 N_1)$ in which $N_1$ represents the number of pixels of each image and $c_1$ the number of (constant) operations required to perform each transformation. Further, as transformation applied to each pixel is independent of others, this step lends itself to the most obvious data-parallelization.

In our next step, we perform K-Means clustering on the 'L' values obtained in the previous step. We first show the time-complexity derivation for the naïve K-Means implementation, and then go on to show how our fast implementation has been able to reduce the complexity. Let '$t_{dist}$' be the time to calculate the distance between any two feature vectors on which we are running the K-Means algorithm; then, each iteration of K-Means has the complexity $O(KN_2 t_{dist})$, where $K$ denotes the number of cluster centres and $N_2$ the number of feature vectors ('L' values, in our case). For finding '$t_{dist}$', we need only one computational step, that of finding the absolute difference between two integers falling within the range [**0**, **255**] (the range of possible values for 'L'). More importantly, since we perform clustering on the bins of the histogram of the left image, and there can be a maximum of **256** bins (representing the range [**0**, **255**]), this drastically reduces the time complexity compared to the naïve implementation which would have run each iteration on all $N_1$ 'L' values (one value for each of the $N_1$ pixels of the left image).

The next three steps, namely, segment boundary detection and refinement using morphological filters 'fill' and 'remove' are performed in number of comparisons linear in the number of pixels of the left image, i.e., $N_1$. Hence, they have time complexities of $O(c_2 N_1)$, $O(c_3 N_1)$ and $O(c_4 N_1)$ respectively with the constant terms representing the (constant) number of operations required for each step. It is also evident, that since the operations for each step are independent for each pixel, they are ideal candidates for data-parallelization leading to accelerated processing.

The next step, viz. finding connected components has three basic operations:

1. Search for the next unlabelled pixel, $p$.
2. Use a flood-fill algorithm to label all the pixels in the connected component containing $p$.
3. Repeat steps **1** and **2** until all the pixels are labelled.

It is clear that searching for new unlabelled pixels until all image pixels have been labelled requires a number of operations which is linear in the total number of image pixels. Further, a recursive flood-fill algorithm works as outlined below:

1. If the pixel we want to label is unlabelled, then label it, else stop.
2. Reclusively flood-fill each unlabelled 8-connected pixel of the above pixel.

Since there are 'if' clauses, flood-fill only considers unlabelled pixels, which have a static number, say, $n$. Therefore worst-case complexity becomes $O(n)$. For ordering $N_c$ connected components based on their constituent number of pixels, we can use merge-sort to finish the operation in $O(N_c \, log(N_c))$ comparisons.

Next, to find the disparity of each boundary pixel using the SAD cost function for a (constant) window size '$w$' and a (constant) maximum disparity '$d$', we require $O(w^2d)$ operations. Now, for $N_b$ boundary pixels, it becomes $O(N_b w^2 d)$. Further, since the disparity of boundary pixels can be calculated independently, they also prove to be ideal candidates for data-parallelization for faster processing.

Lastly, both the 'fill' and 'peek' stages of the depth map reconstruction phase of our algorithm are primarily based on a (constant number of) computations on the disparities of boundary pixels with time complexities of $O(N_b)$ each, and these can also be converted into data-parallel workloads due to its independent nature.

Space complexity of our proposed algorithm can be computed as follows: the pixels' 'L' values and the complete depth map need $O(N_1)$ memory; we also need an auxiliary storage of $O(N_b)$ size for holding the intermediate results of segment boundary detection and refinement and the disparities of pixels lying on refined segment boundaries. For K-Means clustering, an array of size $O(256)$ is required for storing the count of image pixels mapped to each histogram bin for the present iteration (each histogram bin represents each grey-intensity present in the image).

***Parallel Implementation and its Time Complexity Analysis:***

We carry out parallelization of some of the computationally intensive steps of the proposed algorithm whose constituent operations are mutually independent. By executing these independent steps on multiple cores of the CPU and GPU we can reduce the time complexity. We use the Java Thread Pool (JTP) for CPU-parallelization and APARAPI [11], an API which allows suitable data parallel Java code to be executed on GPU via OpenCL, for GPU-parallelization. We also experiment with combining the two methods to achieve even greater performance.

The independent execution units in Java Thread Pool are the "worker threads" and in APARAPI are the "kernels". Thus, we put the parts of our code which we want to run in parallel in the respective execution units. Moreover, combination of CPU and GPU parallelization leads to sharing the workload between them, so we have to decide on how many execution units we want to run simultaneously on the CPU and the GPU, so as to achieve optimal balancing of workload between them.

Thus, assuming the availability of '*p′*' parallel processing units, for four steps of our algorithm, viz. colour space conversion and segment boundary detection and refinement using the morphological filters 'fill' and 'remove', each of them can be completed in $O(c_i N_l/p)$ time. Similarly, the disparities of all boundary pixels can be calculated in $O(N_b w^2 d/p)$ time, and both the 'fill' and 'peek' steps of the depth map reconstruction phase can be completed in $O(N_b/p)$ time each.

## Results and Discussion

We present the outputs of each step of our proposed method to demonstrate how our algorithm works for parameter values given in Table 3.

*Colour Space Conversion:*

*Process:* Left and right input images are converted from the RGB to the CIE-Lab color space following the pixel value transformations outlined in [9].

*Results:* Each pixel of both images has an L value for its lightness component. Fig. 3 (left) shows the converted left input image.

*Observations:* The output image shows the lightness value of each pixel.

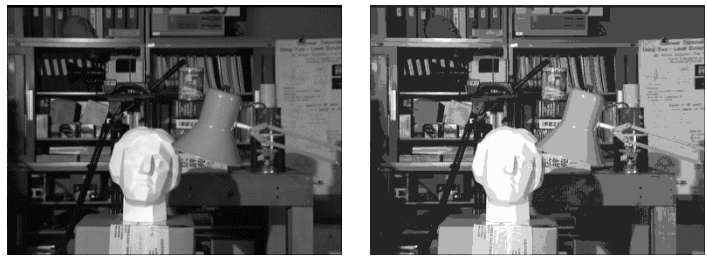

**Fig. 3**  Segmentation of 'L' Values of Left Image using K-Means

*Segmentation:*

*Process:* Converted left input image image is fed to the fast implementation of K-Means clustering algorithm as discussed earlier.

*Results:* The result is shown in Fig. 3 (right).

*Observations:* Small variations in 'L' values (noise) have been abosorbed into single coherent segments representing image regions of pixels belonging to the same object, with similar values of disparity. However, since we have segmented the left image based on just the pixels' 'L' values, ignoring their spatial locations (to expedite the clustering process), we must refine the boundaries of objects defined by the clusters above, so that the spatial distribution of the 'L' values are also taken into account.

*Segment Boundary Detection and Refinement:*

*Process:* We detect and refine the segment boundaries using the approach desribed in the previous section.

*Results:* The outputs are shown in Fig. 4, with output of segment boundary point detection, followed by that of morphological filtering, and lastly, output of connected components analysis at the bottom.

*Observations:* We can infer from Fig. 4b, how the redundancies in segment boundary detection have been removed from Fig. 4a, and from Fig. 4c, how the small artefacts in Fig. 4b have been removed by the connected components analysis technique.

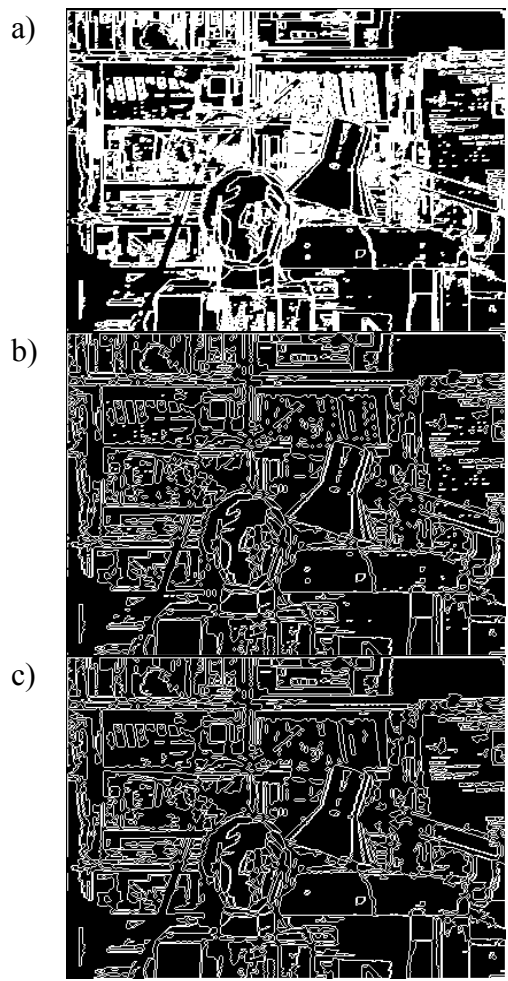

**Fig. 4** Segment Boundary Detection and Refinement

For the Tsukuba image pair, the segment boundary refinement reduces the number of boundary pixels by nearly 53% for the parameter values of Table 3, such that only 19% of the number of pixels of the left image are used for disparity calculations. This greatly reduces the number of disparity computations in the next step.

***Disparity Measurement of Boundary Pixels:***

*Process:* Disparities of the boundary pixels are determined by using the SAD algorithm.

*Results:* The result in shown in Fig. 5 (left).

*Observations:* Boundaries of objects closeer to the camera (like the lamp and the head of the statue) are having a higher intensity (greater value of stereo disparity). This is in direct agreement with the reality that value of disparity of a pixel is inversely proportional to its distance (from the camera lens / human eyes).

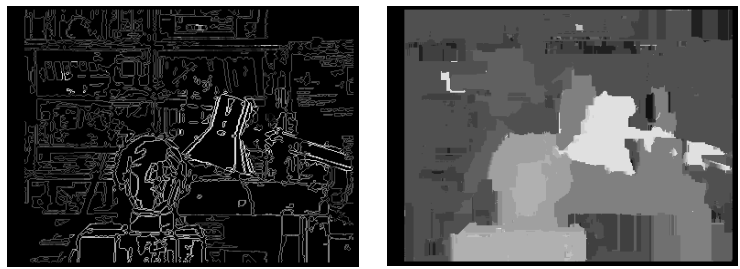

**Fig. 5** Disparity Map Reconstruction from Boundary Disparities

***Disparity Map Reconstruction from Boundaries:***

*Process:* Disparities of the segmented regions in the image are determined from the disparities of their boundaries using the two-stage disparity map reconstruction method discussed earlier.

*Results:* The results are shown in Fig. 5 (right).

*Observations:* The depth map shows the mapping of each image point to a certain level of disparity. Objects which are closer to the camera like the lamp-shade have a greater disparity value resulting in a brighter shade of gray. For objects further away from the camera like the table, the shades of gray get progressively darker, as their disparity values keep on decreasing.

To evaluate the disparity estimation of our approach vis-a-vis those of existing algorithms, we chose three pairs (left and right) of images from the Middlebury stereo vision data-set, viz. Tsukuba, Sawtooth and Venus. We ran our code on those three image pairs. We then compared the depth maps obtained by our algorithm against the corresponding ground truth depth maps provided in the dataset, as well as with the depth maps created by two established methods (SAD and NCC [2]) and a recent one [1] for those same three image pairs. We present the results of those comparisons below.

**Comparison with Ground Truth Depth Maps:**

To evaluate the depth estimates of our proposed method, we consider the approach of computing error statistics with respect to the ground truth image available with the Middlebury dataset, and we choose the quality metric as the percentage of bad matching pixels (B) given by Eq. 3,

$$B = \frac{1}{N} \sum_{(x,y)} (|d_C(x, y) - d_T(x, y)| > \delta_d) \qquad (3)$$

where $d_C(x, y)$ are the computed disparities and $d_T(x, y)$ are ground truth disparities, and $\delta_d$ denotes the disparity error tolerance, which is taken as '1.0' as per published works [8].

Table 3 shows the values of our algorithm's parameters which gave best results are hence, were used for performance comparison. From the table, it's evident that the set of optimal parameter values is almost constant across all three images. We have observed in our experiments that slight variations of the parameter values about the optimal ones effect negligible changes in accuracy of depth estimation.

**Table 3** Parameter Values for the Three Image Pairs

| *Parameter* | *Tsukuba* | *Sawtooth* | *Venus* |
|---|---|---|---|
| Number of clusters (K) for K-Means | 10 | 10 | 10 |
| Block size for cost aggregation (odd) | 9 x 9 pixels | 9 x 9 pixels | 9 x 9 pixels |
| Disparity threshold for reconstruction | 0 | 1 | 1 |

Fig. 6 presents a comparison of the depth map produced by our method (left) with ground truth depth map (right) for the Tsukuba image. The black regions in depth maps denote regions whose disparities are not compared. Further, the percentage of bad matching pixels was calculated as 7.8%.

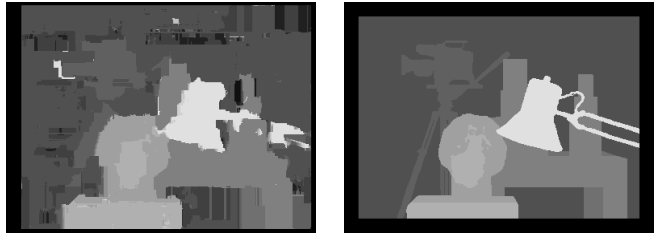

**Fig. 6** Resultant Depth Map Compared to Middlebury's Ground Truth

It can be seen from the comparison of our depth map with the ground truth that our algorithm yields fairly accurate depth estimation; its performance is affected mostly in some texture-less and low illumination regions. But as we will be using the generated depth maps for the purpose of selective blurring, inaccurate depth estimates in regions of low texture and illumination will not affect the blurring performance, as human eyes are more sensitive to variations in brightness than colour, and hence perceptual differences resulting from blurring a dark region would be negligible. Likewise, as blurring is performed using a weighted average of neighbouring pixel values, a region of low texture will remain relatively unaffected after a blurring operation, as the variation in pixel values in a texture-less region is negligible. Some incorrect depth estimations also occur in occluded regions which can be reduced using occlusion handling techniques in future. Also, our algorithm does not calculate depth information of pixels lying near the image borders (which we will address in our future work), but this limitation does not significantly affect the output of depth-based blurring, as the images' borders very rarely contain objects or regions of interest to the user.

**Comparison with other Depth Estimation Methods:**

Table 4 shows the results of a performance comparison of our depth detection method with two established methods, viz. Sum of absolute differences (SAD) and normalized cross-correlation (NCC) (both of which have been used in numerous traditional algorithms for stereo depth estimation), as well as a recent work [1] which uses absolute difference (AD) cost function to predict disparities. Again, we use the percentage of bad matching pixels to quantitatively compare depth maps generated by our method, SAD, NCC and [1].

Table 4  Performance Comparison of Proposed Algorithm

|  | *Tsukuba %* | *Sawtooth %* | *Venus %* |
|---|---|---|---|
| NCC [2] | 41.4 | 9.92 | 17.4 |
| SAD [2] | 36.9 | 11.9 | 24.5 |
| Zhen Zhang et al. [1] | 13.9 | 7.22 | 6.12 |
| Proposed Algorithm | 7.8 | 5.26 | 4.72 |

The results clearly demonstrate that our proposed method outperforms traditional NCC and SAD methods by up to 33.6%, and even recent method [1] by up to 6.1%.

Further, the percentage of left image pixels used for stereo depth estimation was about 19% for Tsukuba, 18% for Sawtooth and 16% for Venus image pairs, implying that our algorithm is scalable to high resolution stereo images.

Figs. 7 & 8 show depth maps generated by our method (left) and ground truth depths (right) for the Sawtooth and Venus images. All other depth maps supporting performance comparison data presented in Table 4 can be found in [1].

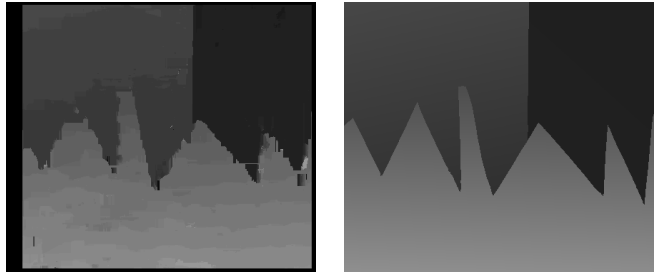

**Fig. 7**  Sawtooth Depth Map for Proposed Algorithm (left) and Middlebury's Ground Truth (right)

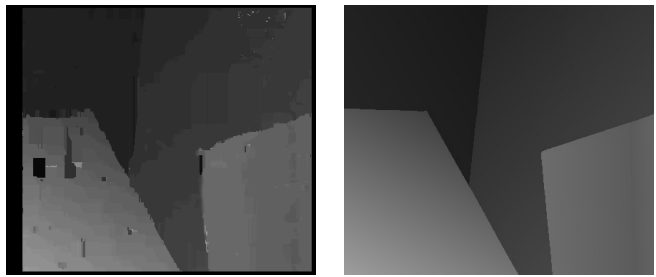

**Fig. 8**  Venus Depth Map for Proposed Algorithm (left) and Middlebury's Ground Truth (right)

Next, we present the results of depth-based blurring on the Tsukuba image using depth maps output by our proposed algorithm in Fig. 9. In the top-right image, only the lamp, and in the bottom-left one, only the statue-head and the table are in focus, as per input depths ranges of interest. Multiple in-focus depth ranges, like the bottom-right image extend the methods like [10]. It is interesting to note that the incorrect depth estimates of our algorithm in the texture-less and low illumination regions do not (noticeably) affect the depth-based blurring.

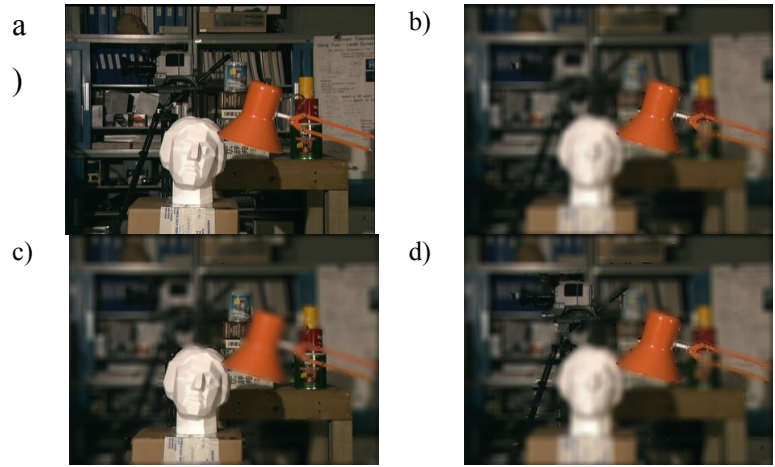

**Fig. 9** Original Image (a), Depth-based Blurring with Single Depth Range in Focus (b & c) and Depth-based Blurring with Multiple Depth Ranges in Focus (d)

Lastly, we present the results of parallelization using CPU (Java Thread Pool) and GPU (APARAPI) separately, and then by combining them, in Table 5. The experiments were performed on a PC having a Intel $i$7 quad-core CPU with 4 GB RAM and a NVIDIA NVS 300 GPU having 2 compute units and 512 MB RAM. In the "combined" implementation, we chose to run four JTP worker threads and two APARAPI kernels simultaneously, as this gave us best results.

**Table 5** Results of Parallelization of Proposed Algorithm

| No. of Images | Serial Time (ms) | JTP-only Time (ms) | APARAPI-only Time (ms) | JTP+APRARAPI Time (ms) |
|---|---|---|---|---|
| **150** | 73191 | 32396 | 17331 | 13822 |
| **200** | 96361 | 42391 | 22982 | 16595 |
| **250** | 116909 | 47070 | 28514 | 20192 |

The increase in processing speed obtained by parallelizing serial algorithms and running on CPUs / GPUs is quantitatively described by a metric called speed-up factor as shown in Eq. 4:

Speed-up = ExecutionTime$_{SERIAL}$ / ExecutionTime$_{PARALLEL}$ (4)

Speed-ups achieved over serial implementation by parallelizing our proposed algorithm and executing the serial and parallel versions on the CPU and GPU with a input stereo image sequence of 4096x2304 pixels each are given in Table 6. The maximum speed-up obtained is 5.8 for a sequence of 250 stereo images, for the CPU+GPU (combined) parallelization approach. It has been seen that, generally with the increase in the number of multi-processing units (cores) and the increase in the speed of each processing unit, the parallel executions become quicker and hence the speed-up increases following a sub-linear trend. It is not perfectly linear due to a number of causes, the most significant of which is a factor called parallelization overhead, which deals with coordinating activities of different processing units to collectively achieve a single task, like the distribution of the total workload among all parallel processing units, collecting the intermediate processing results from each unit and compiling them to form the output.

Table 6  Speed-ups Achieved by Parallelization of Proposed Algorithm

| *No. of Images* | *JTP-only Speed-up* | *APARAPI-only Speed-up* | *JTP+APRARAPI Speed-up* |
|---|---|---|---|
| **150** | 2.3 | 4.2 | 5.3 |
| **200** | 2.3 | 4.2 | 5.8 |
| **250** | 2.5 | 4.1 | 5.8 |

From the graph of Fig. 10, we can easily infer that the rate of increase of running time of our proposed algorithm is greatest for the serial implementation, followed successively by the JTP, APARAPI and the JTP+APRARAPI (combined) ones, in decreasing order. This is because GPUs are more suitable for running data-parallel workloads than CPUs, and the combined approach utilizes both for parallelization.

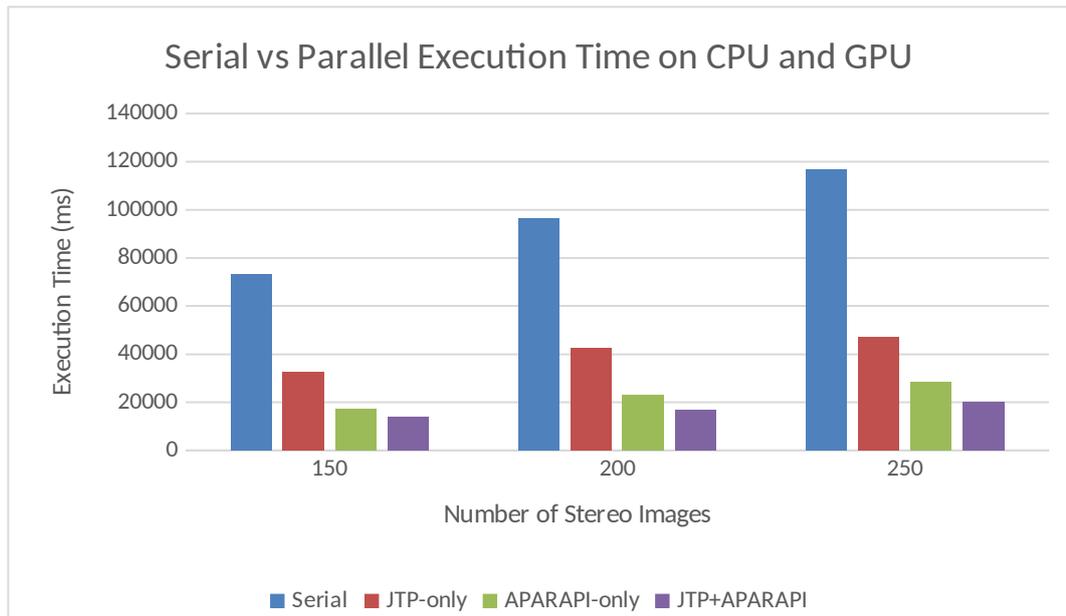

**Fig. 10** Results of Parallelization of Proposed Algorithm

## Conclusion and Future Work

In this paper, we proposed a stereo depth estimation method using only 18% pixels of either the left or the right image, which outperforms traditional methods like SAD and NCC by up to 33.6% and a recent method [1] by up to 6.1%. We also performed depth-based Gaussian blurring of image regions as per depths of users' non-interest. We showed that despite our algorithm's inaccurate depth estimates in texture-less and low-illumination regions, its blurring performance is not affected. Future work will focus on improving the depth estimates and the depth-map reconstruction technique.